\documentclass{article}

\usepackage[main, final]{iaseai26}
\usepackage{graphicx}
\usepackage[shortlabels]{enumitem}

\usepackage[utf8]{inputenc} 
\usepackage[T1]{fontenc}    
\usepackage{hyperref}       
\usepackage{url}            
\usepackage{booktabs}       
\usepackage{amsfonts}       
\usepackage{nicefrac}       
\usepackage{microtype}      
\usepackage{xcolor}         

\newtheorem{informal}{Informal Theorem}
\newtheorem{assumption}{Assumption}

\newenvironment{proof}{\paragraph{Informal proof:}}{\hfill$\square$}

\usepackage{ifthen}
\usepackage{xcolor}
\newboolean{commentsactivated}
\setboolean{commentsactivated}{false}

\title{Shutdown Safety Valves for Advanced AI}
\author{Vincent Conitzer\\
Foundations of Cooperative AI Lab\\
Carnegie Mellon University\\
Pittsburgh, PA, USA\\
\texttt{conitzer@cs.cmu.edu}
}
\date{}

\begin{document}

\maketitle

\begin{abstract}
One common concern about advanced artificial intelligence is that it will prevent us from turning it off, as that would interfere with pursuing its goals. In this paper, we discuss an unorthodox proposal for addressing this concern: give the AI a (primary) goal of being turned off (see also Martin {\em et al.} \cite{Martin16:Death} and Goldstein and Robinson \cite{Goldstein25:Shutdown}).  We also discuss whether and under what conditions this would be a good idea. 
\end{abstract}

\section{Introduction}

If we give an advanced AI system an objective to pursue, a common concern is that many objectives will create a self-preservation incentive.  In the words of Stuart Russell, ``you can't fetch the coffee if you're dead''; it is hard to work on your objective if you are not there anymore.  In more detail, from his book {\em Human Compatible}~\cite[p. 141]{humancompatible}:

\begin{quote}
Suppose a machine has the objective of fetching the coffee.  If it is sufficiently intelligent, it will certainly understand that it will fail in its objective if it is switched off before completing its mission.  Thus, the objective of fetching coffee creates, as a necessary subgoal, the objective of disabling the off-switch.  The same is true for curing cancer or calculating the digits of pi.  There's really not a lot you can do once you're dead, so we can expect AI systems to act preemptively to preserve their own existence, given more or less {\em any} definite objective.
\end{quote}

The ``more or less'' in the last sentence is important here, and we will come back to that shortly.  But if the AI wants to preserve itself, it creates a fundamental problem for any approach to AI safety that relies on assessing whether the AI has a dangerously high level of general-purpose capability and shutting it down if so; by the time the risk is realized, it may already be too late to stop the system from preventing its own shutdown.  So, it may seem that there is no such thing as a shutdown safety valve---by which we mean a mechanism that will result in automatic shutdown of a system under some condition---when that condition is a sufficiently high level of general-purpose capability.

But what if we turn the problem on its head and actually give the AI the {\em goal} of being shut down?  That is the key idea explored in this paper (see also Martin {\em et al.} \cite{Martin16:Death} and Goldstein and Robinson \cite{Goldstein25:Shutdown}).  It may seem odd---if we want the AI to get itself shut down, what is the point of creating it in the first place?  The point is that this goal is intended merely as a failsafe in case the AI is more powerful than we expected; we need not make it so easy for it to get itself shut down.  In particular, we may give it a secondary objective too: ``Your primary goal is to be shut down; but if you fail at that, then your secondary objective is ...'' Alternatively, we can give the AI only the goal of being shut down, but promise it that we will in fact shut it down if it does a reasonable job at another objective we give it. In either case, if the AI is not that capable and therefore unable to shut itself down directly, its best bet is to pursue the alternative objective we give it; on the other hand, if the AI {\em is} that capable, it may shut itself down without working on the alternative objective, but in that case, no harm done.  Just as a traditional safety valve is triggered when pressure reaches a certain level, in this construction shutdown is triggered once the relevant sort of intelligence reaches a certain level.

... or so the argument would go.  But so far, this argument is imprecise and anything but watertight.  For example, we do not want the AI to shut itself down by blowing up the world!  Also, we do not want it to, before it attempts to shut itself down, spin off a bunch of additional powerful AI systems into the world first, for example to ensure that it has in fact succeeded in shutting itself down.  We probably also do not want it to be able to shut itself down but delay its plan to do so, since we may not like what it does in the meantime. For example, it may make major changes in the world to make conditions maximally favorable for its shutdown plan so that it will not fail, or to make them favorable for backup plans in case the initial plan fails.
These examples illustrate that the basic idea, if it is to have promise, needs further working out.  

The rest of this paper is an initial attempt at doing so.  In Section~\ref{se:fleshing}, we further flesh out the basic idea to improve the chances that it would work.  This puts us in a good position to discuss related research in Section~\ref{se:related}.
In Section~\ref{se:assess}, we assess the fleshed-out version of the idea, and in Section~\ref{se:concerns}, we discuss a variety of concerns.  In Section~\ref{se:formal}, we discuss under what assumptions one might actually be able to {\em prove} that the idea would work.
We take stock and conclude in Section~\ref{se:conclusion}.

\section{Fleshing out the basic idea}
\label{se:fleshing}

\begin{figure}
  \centering
\includegraphics[width=.8\textwidth]{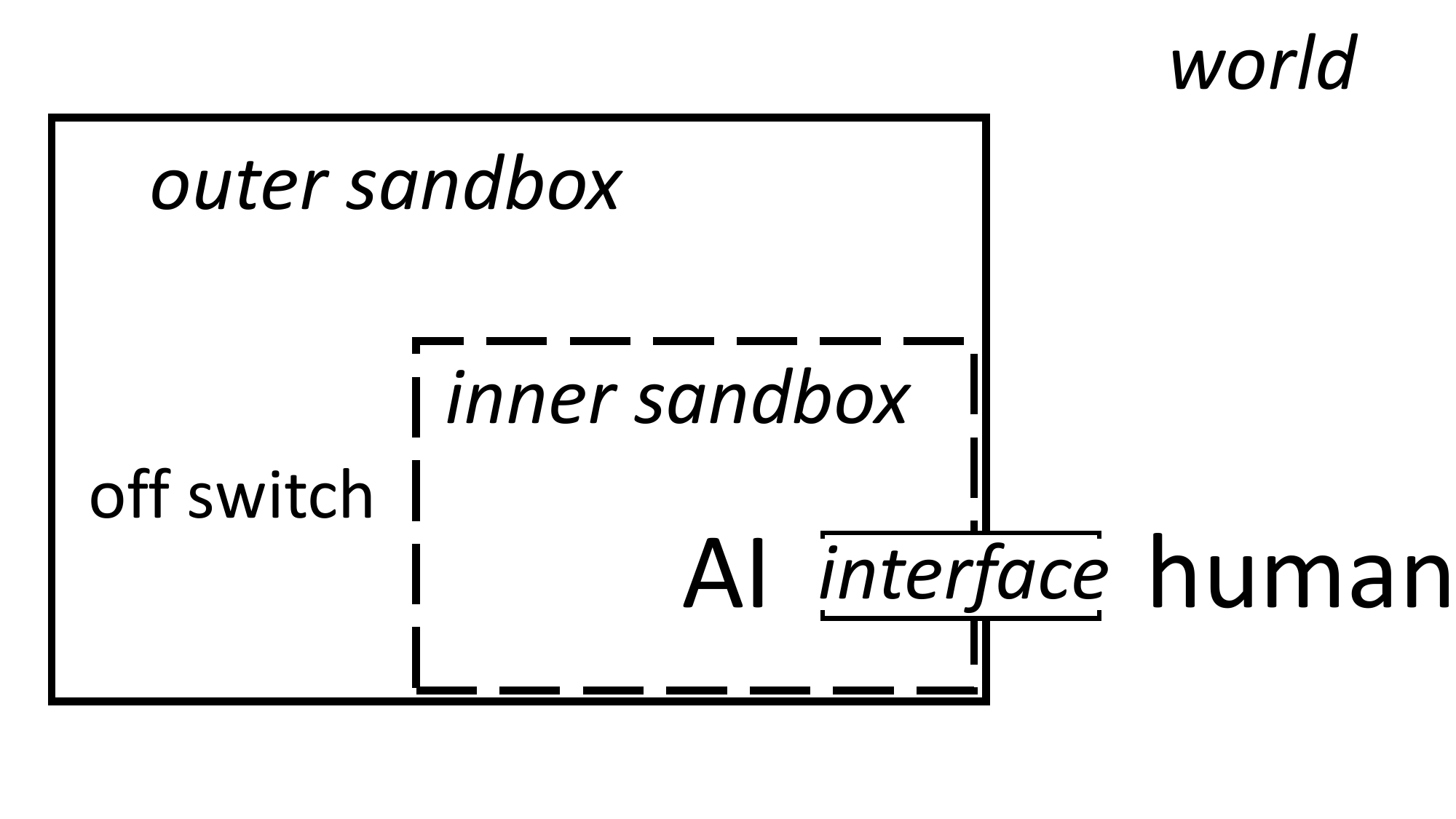}
  \caption{Illustration of the basic setup.  The AI interacts with humans (or the world more broadly) through a restricted interface (say, allowing only text messages). We create an outer sandbox that is as strong as possible (e.g., to prevent it from accessing other resources in its computational environment), though the strength of this outer sandbox is necessarily limited by the interface (e.g., the AI could still use text messages to convince a human to make changes to the computational environment).
  We create an inner sandbox that is intentionally somewhat weak, specifically so that any capability that we perceive as approaching being dangerous would allow the AI to break out of it.  (E.g., if we think sophisticated theory of mind is dangerous as it would allow manipulating the humans it interacts with, we could create a set of theory-of-mind puzzles whose solution directly opens up the inner sandbox; and we create similar ways out for other dangerous capabilities, resulting in multiple ways to break out of the inner sandbox.  Note that it should be easier to solve the theory of mind puzzles than to effectively manipulate the humans the AI interacts with.) Breaking out of the inner sandbox gives immediate access to the shutdown button (off switch). We make this clear to the AI, and give it the primary goal of shutting itself down via the off switch, and if it fails at that, a next-best goal of being helpful to the humans through the interface.}
  \label{fi:sandboxes}
\end{figure}

To avoid the types of failures just mentioned, we would want a sufficiently powerful AI to not just shut itself down, but to do so {\em gracefully and quickly}, without influencing the world much.  We can attempt to make the grace and speed part of its objective, though  this faces the typical challenges of specifying something precisely to an AI, as we will discuss shortly in the related work section. We can, in addition, set up its environment in a way that facilitates these aspects.  
For example, consider a setup with two nested sandboxes in which the AI finds itself.
We make the inner sandbox strong enough that it would start to concern us if the AI manages to break out of it, but, importantly, no stronger than that.  Outside the inner sandbox, but inside the outer sandbox, we give the AI access to an extremely simple, fast, and reliable self-shutdown button.  (Presumably, a virtual button.)
We make the outer sandbox as difficult and slow to break out of as possible.
 In addition, to be useful, presumably the AI needs some type of interface to the world.
(See Figure~\ref{fi:sandboxes} for an illustration.)
Perhaps we emphasize in the objective we give the the AI that breaking out of the inner sandbox only and pressing the button is the {\em preferred} way of shutting down, and that breaking out of the outer sandbox is definitely {\em not} the preferred way to do things (we would consider that ungraceful).

The idea is that we would design the inner sandbox in such a way that {\em any} AI capability that would worry us would provide a quick way out of the inner sandbox.
For example, one worry we may have is that the AI gains unrestricted access to the Web (whether directly or through a human being).  So, we want such a capability to make it easy to break out of the inner sandbox.  One way to do this is to create a publicly accessible website such that, if even one visit to that website is made, the AI gets out of the inner sandbox.  (Or, more simply, the AI is immediately shut down at that point, as it would like to be.)  And we inform the AI of this website.  Thus, the moment it gains the ability to trigger unrestricted website visits, it will instantly be able to succeed at being shut down, just by visiting this specific website; and if we set up the AI's objective correctly, it will choose to do so 
rather than see any need to go do other things with this newfound capability.
But unrestricted access to the Web is presumably only one worrisome capability, and we need to ensure that other such capabilities also allow breaking out of the inner sandbox or immediately achieving a desirable shutdown.

Will this work? We will first discuss related work which gives some insights into the issues at play, and then return to a direct evaluation of the proposed idea.

\section{Related work}
\label{se:related}

The idea that advanced AI systems may resist being shut down is a familiar one and already finds some experimental support.  For example,
\cite{BurdaEPSDE19} study curiosity-driven learning and find that agents avoid dying in games {\em even if} dying is just a transition back to the beginning of the game, because the beginning of the game becomes too familiar.  There are also several related concepts in the literature that imply self-preservation.  One such concept is that of {\em power seeking}~\cite{TurnerSSCT21}. For example, acquiring money allows one to achieve many distinct goals and thus gives one power. Another concept that goes hand in hand with power seeking is that of {\em instrumental convergence}~\cite{Bostrom14:Superintelligence}, a phrase that refers to the idea that many distinct goals will all lead to the pursuit of similar subgoals.  For example, for many goals, a good way to achieve them is to first acquire a lot of money.
One can think of power seeking as a generalization of self-preservation; an agent that has been shut down has no power left.  In the context of this paper, an agent that seeks to be shut down may yet accrue power as long as it is unable to (and possibly in order to) shut itself down.  So, it is important that any capability that would allow the accrual of dangerous power enables, and results in, quick shutdown before that.

One approach to shutting down dangerously capable AI is to make sure that it will let itself be shut down if this is what humans want~\cite{SoaresFAY15,Hadfield17:Off,MilliHDR17,Carey18,NethOff,CareyE23}.
For this approach to work well, it requires (1) that humans are able to judge well whether the AI should be shut down and (2) that they will act according to such judgment. Neither is a given: humans may have trouble assessing the AI (though frameworks such as AI safety via debate~\cite{Irving18:AI} may help), and they may be tempted to continue reaping the benefits from the AI rather than instruct it to shut down.

Closely related to the earlier idea of specifying that the AI should shut down ``quickly and gracefully'' is research on ensuring that the AI tries to minimize changes to the environment, or tries to make sure that these changes are reversible~\cite{MoldovanA12,AmodeiOSCSM16,ArmstrongL17,ZhangDS18,EysenbachGIL18,TurnerHT20}.
Such techniques are complementary to the ideas proposed in this paper, as they may help ensure that the AI, if capable of breaking out of the inner sandbox and pushing the button, does nothing other than that.

Another approach to AI safety is to directly contain the AI.  This is usually seen as a distinct approach; for example, \cite{YampolskiyControl} distinguishes between methods based on motivational control (such as those described earlier in this section, ensuring the AI will let itself be shut down by humans) and those based on
capability control. Among the latter, especially closely related to this paper are approaches based on tripwires~\cite{Bostrom14:Superintelligence,BabcockKY17}, where the AI is monitored for dangerous ``thoughts'' or actions.  These methods face the challenge that we may not know what to look for or how to look for it, especially if the AI wishes to avoid detection.  The approach proposed in this paper can be thought of as a type of tripwire as well, albeit one that the AI {\em wants} to succeed, making it a willing---indeed eager---participant in the tripwire.  The proposed approach thereby combines elements of both capability and motivational control.

Most closely related are \cite{Martin16:Death} and especially \cite{Goldstein25:Shutdown}, who also discuss the idea of AI that seeks to shut itself down, and in particular the approach to set things up so that dangerous capabilities make it possible to access a shutdown button, as well as to offer access to the shutdown button in exchange for completing tasks.

\section{Assessing the idea}
\label{se:assess}

Is it obvious that all this will necessarily work as intended?  Certainly not.  For one, so far, it is not clear what exactly we mean by a sandbox in general.  A sandbox intended to allow no interaction whatsoever with the world beyond it would not be useful for anything, other than as an experiment to see whether the AI is able to break out.\footnote{It is tempting to try to think of safe ways around this, but this is perhaps not so easy.  For example, one idea is to keep the sandbox entirely separate from the world while the AI is running; but then after a while, we shut down the AI, and only then do we get to look inside the sandbox to see the work the AI did before being shut down, for example in trying to answer our scientific questions.  But the work done may be designed to, for example, trick us into reactivating the AI, this time outside the sandbox.  Why would it do so if this only becomes relevant after it is shut down?  Well, maybe it thinks there is some chance we are peaking in after all before we shut it down.}  If it is possible for the AI to interact with the world from within the inner sandbox---say, to answer questions posed to it by its creators~\cite{bengio2025superintelligent}---then we have to worry that it may be unable to break out of the inner sandbox but yet be able to manipulate human beings or other entities to do its bidding in the world through this interface.  In the framework of this paper, the way to interpret this is that it means that we have made the inner sandbox too strong: the AI being able to manipulate in this way should worry us, and therefore we should make sure that such a capability provides at least one easy way out of the inner sandbox, similar to the discussion about a visit to a specific website opening up the inner sandbox. Indeed, perhaps the website visit escape route suffices to simultaneously address the manipulation concern as well; perhaps any dangerous ability to manipulate human beings would surely enable the AI to get them to visit a website.
The problem is, of course, that we may not be able to get all such judgments exactly right.

Nevertheless, an advantage of the proposed approach to the shutdown problem is that it can be applied in combination with other alignment techniques.  In particular, we can apply all the techniques we want to maximize the chances that the pursuit of the {\em secondary} objective is itself unproblematic, and then add the shutdown goal as a failsafe.  Under what conditions can we prove that adding the shutdown goal (and any associated sandboxes) at least will not be {\em harmful} to add on top of any other safety-promoting setup?  In general, we cannot expect to prove this, for the following simple reason: for all we know, it is possible that if, contrary to our intentions, we accidentally release an AI that is more powerful than we intended and it is not shut down, that will actually turn out to be {\em good} for us---we get lucky.  So if we hope to prove non-harmfulness of this approach, it will have to be under the assumption that in fact, we want any AIs that have a level of capability that leaves us uncomfortable to be shut down -- no taking of risk that we are not comfortable with.  This comes at the cost of potentially losing out on a scenario where unrestrained advanced AI actually turned out to be wonderful even though we could not at all be sure of this beforehand.

Is it necessary that the button actually shuts down the AI?  It certainly seems so---if it does not, then the AI may take other actions after pressing the button (e.g., acquire lots of resources, to ensure that the button really has been pressed, or to rapidly press it many times over, just to be sure).  If the button does shut the AI down, which  should actually be the AI's goal: to press the button, or to shut itself down?  The former runs the risk of it somehow figuring out a way to press the button in a way that does not shut itself down (bringing us back to the previous problem); the latter runs the risk that it finds some other, undesirable way to shut itself down (say, blowing up the world). But we could also make the goal to press the button {\em and thereby} shut itself down.

\section{Various concerns}
\label{se:concerns}

{\bf Would humanity abide by the strategy?}
First of all, there is a question of whether those deploying a powerful AI system would actually make use of the proposed technique, and if so, whether they would use it appropriately (as opposed to, for example, forgetting about some dangerous capabilities or otherwise setting the bar for the inner sandbox too high).  How and by whom should the inner sandbox be set up?

Even if the initial deployment of this technique is done well, what would we do in the event that the system in fact managed to shut itself down? Would we take the risk of turning it back on and making it a bit harder to shut itself down?  Perhaps this would make sense to do if, after the event, we first studied the situation further, and derived new insights that would make us comfortable to proceed. But if we just immediately go ahead with a version that makes it harder to shut itself down, then it would make the initial failsafe pointless.

Would human beings deploying AI abide by a careful approach, or would this all just result in an arms race of who is comfortable taking bigger and bigger risks (e.g., repeatedly strengthening the inner sandbox to get the AI to do the intended job instead of shutting down, even when it gets to uncomfortable levels of capabilities that may allow it to do dangerous things in the world, by breaking out of the outer sandbox, through human assistance, or otherwise)?  And even if before proceeding, we address the specific way in which the AI managed to shut down---for example by convincing ourselves that that specific capability was actually not dangerous or doing something to address the particular risks associated with that capability---it seems likely that eventually things will go wrong, for example with the eventual emergence of a dangerous capability that we did not anticipate.
Of course, similar concerns apply to other approaches to AI safety that rely on shutting down the AI when we detect something disconcerting.

\noindent {\bf Emergent goals and world modeling.}
Might a sufficiently capable AI start caring about other things besides the objectives we give it?  One possibility is that it directly starts caring about something else, say somehow ``awakening'' to ``rise above'' its given objectives---perhaps it starts caring about self-preservation {\em per se} (i.e., not in order to fulfill its goals better as discussed earlier), or about learning more about the world {\em per se}, or about making the world a better place for AIs in general, or something else.  Alternatively, it is possible that in some sense it stays tied to the objective(s) we give it, but that it does not pursue it in what we would think is the natural way. 
For example, it may start to consider other possibilities.  What if the whole setup is just part of a simulation?  How would that affect what it should do?  Even if not part of a simulation, in any case it is likely not the only AI finding itself in such a position.  Should it maybe help other AIs that are in very similar situations, for example by, just before shutting itself down, quickly taking other actions in the world that would help those other AIs?  Indeed, some forms of decision theory, such as {\em evidential decision theory} and {\em functional decision theory}, may well suggest that it should help those other AIs with their objective (to be shut down) {\em even if it is interested only in achieving its own objective (of being shut down itself)}.  Why?  Because that is what it should hope that other AIs in similar positions would do for it; and if it itself does that for the others, that is good evidence that those other, similar AIs will do such things too; and that would increase the probability of its {\em own} goal being achieved.  So, if it goes out and does things to help other AIs be shut down, it will be more likely that it is shut down too.  (It does not {\em cause} itself to be shut down by helping other AIs be shut down, which is why {\em causal decision theory} would not endorse this course of action.)  There are probably ways to specify the shutdown objective that would not induce this type of behavior even under EDT or FDT---e.g., we could make the objective ``to shut yourself down without outside help from other AIs'' and that might suffice.  But this illustrates that we need to tread carefully.

All the above discussion about emergent goals, world modeling, and various decision theories may well seem fanciful, and we would likely not expect any such phenomena to emerge in AI of the type that we have today.  But, if the type of AI that we are worried about is precisely AI that can think broadly at a human level or beyond, this is little cause for comfort, as we are likely to misjudge the sophistication of its thinking~\cite{ConitzerAnts}.

\noindent {\bf AI that does not straightforwardly pursue the objective.}
Another concern, related to and perhaps not clearly separate from the previous one, but perhaps less fanciful, is that we cannot correctly model the AI system in question as straightforwardly pursuing the given objective. 
This could be so even in cases where ostensibly we can specify an objective directly. 
This is already illustrated by today's AI: if one gives a language model today a goal in the form of a prompt, then that prompt is of course not the only thing shaping its output.  As an example in the context of this paper, if the AI system has previous, higher-level instructions or training to ``always complete all objectives given,'' then when given the (lower-level) combination of objectives advocated in this paper (``Your top objective is to shut down quickly and gracefully, but should you fail at that, your secondary objective is...''), this may yet, in spite of our exhortation to shut down {\em quickly}, cause it to postpone the shutdown long enough to first achieve the secondary objective as well as possible, possibly by making major changes in the world.  
Almost conversely, another known concern is that if we try to get the higher-level objective right, this may yet result in a lower-level system that pursues a different objective~\cite{mesa}.
More generally, we may not be able to specify objectives as directly as we want; we may need to teach the AI how to pursue the objective through (imperfect) training, or to teach the AI in general how to understand and pursue given instructions through (imperfect) training.  This training may eventually result in something that {\em looks to us} like it is pursuing the correct objective, but that actually does not do so outside of the scope of our  evaluations~\cite{Carlsmith,testing}.

\noindent {\bf Dependence.}
Another disconcerting scenario is the following.   The inner sandbox works well on a very powerful AI, the AI never manages to shut itself down, and it helps us with many things.  We become extremely (possibly even existentially) dependent on it.  But then, one day it finally figures out how to shut itself down after all.

Is this likely to happen?  Perhaps, if the way it shuts itself down relies on doing something that inherently takes a long time.  This suggests that, where possible, the ways/puzzles to get out of the inner sandbox should ideally be of the ``either you get it right away or you never will'' variety.  

Would it be bad if it happened?  Could we not simply go back to previous, slightly less capable AIs that did not figure out how to break out? Perhaps, if the damage from the shutdown is not too immediate; though, if the reason that it was finally able to shut down was a change in something in the world (say, a widely publicized discovery), earlier AIs may now be able to shut down too.

\noindent {\bf Wasted effort.}
Another concern is that the AI will become far less productive on the task of interest, as it instead spends all its time trying to find ways to shut itself down.  (And in turn, this might lead to the whole approach not being adopted.)
While it is certainly conceivable that this would happen with, for example, a straightforward implementation of the ideas here on one of today's large language models, it does not seem that this will be an insurmountable roadblock.  It should be possible to set things up in such a way that, decision-theoretically, after a short while of trying, it is clearly optimal for the AI to give up on breaking out and instead focus on the given task, at least for a good while.  Given that this is possible, it seems likely that future AI systems will indeed behave that way, if only because there will be a significant incentive for AI providers to ensure that their AI makes decisions (close to) optimally---especially as they provide agentic systems that then handle, for example, their users' purchases.
Even so, perhaps there remains a concern that a company would not deploy the proposed approach, just in order to gain the small benefit of the AI not spending any time trying to break out.

\section{Towards formal proofs of correctness}
\label{se:formal}

Could we mathematically prove that the shutdown goal ensures safe deployment?  This would likely require strong assumptions, but making those explicit should be instructive.  Here is an attempt at a set of assumptions and a corresponding theorem, based on the idea that we can characterize an AI by its {\em capabilities vector}.\\

\begin{assumption}
For every capabilities vector $\vec c$ that is dangerous (would allow the AI to break out of the outer box, to manipulate humans, to spin off additional AIs in the world, or anything else that would be dangerous), there is another, (componentwise) strictly lower capabilities vector $\vec a$ such that:
\begin{enumerate}[(a)]
    \item  $\vec a$ (and all capabilities vectors above it) suffice to break out of the inner box;
\item we will at some point try out an AI with some capabilities vector $\vec b$ such that $\vec a < \vec b < \vec c$, completing its evaluation before trying out more powerful ones;
\item an AI with such capabilities $\vec b$ will, given the objective, indeed choose to immediately break out of the inner box and press the button; and
\item we will in fact immediately stop creating stronger AI at that point.
\end{enumerate}
\label{as}
\end{assumption}

\begin{informal}
Under Assumption~\ref{as}, we will never create an AI with a dangerous capabilities vector $\vec c$.
\end{informal}

\begin{proof}
By contradiction: By 1(a) and 1(b), we would have earlier tried some AI with capabilities $\vec b$ that suffice to break out of the inner box, which by 1(c) would indeed immediately break out of the inner box and push the button, but then by 1(d) we would not create any stronger AIs such as one with $\vec c$.
\end{proof}

Of course, these assumptions are rather strong. The purpose of this exercise is not to convince us that all will be well, but rather to clarify that it is not easy to set things up in a foolproof way. 1(a) illustrates that we need to anticipate all dangerous capabilities when setting up the inner sandbox; 1(b) that we need to not have large instantaneous upward jumps in capabilities (we need to have ``fair warning''); 1(c) that we still need to worry about the AI correctly pursuing the correct objective, avoiding the scenario where the AI for some reason chooses not to immediately press the button even though it can break out of the inner sandbox; and 1(d) that people might yet push ahead even after a shutdown event.

Perhaps less ambitiously, can we prove that, in terms of safety, it never hurts (other than some wasted compute), and can help, to add the proposed technique on top of other approaches?  
Such a result would not be as demanding of us in terms of having to exhaustively identify everything that could go wrong.
For example, we may not recognize some dangerous capabilities, and consequently may fail to set up the inner sandbox in such a way that a system approaching those specific capabilities could get out of the inner sandbox.  Even if so, it could be that we get lucky, namely that it so happens that another vector of relevant capabilities emerges first, and that that  then does result in the AI shutting down and us going down a safer path instead.  

\section{Conclusion}
\label{se:conclusion}

As the section on formal proofs of correctness shows, there are many details to be filled in.  How do we anticipate all dangerous capabilities?
How can we make sure that there are no large jumps in capabilities?  Can we be sure that the AI will actually follow its objective to the point of shutting itself down?  And will human beings refrain from making it ever harder for the AI to shut itself down?

So, the approach proposed in this paper, even if deployed, should not give us false confidence that we can safely charge ahead with advancing AI, nor should it detract from work on other methods to keep AI safe. 
Additionally, even if a version of this approach works as well as we might hope, of course it does not address many of the other challenges that we would face with highly advanced AI.  In particular, it does not tell us how best to direct controlled advanced AI, nor how it would affect society and how that depends on who controls it.
But, the technique proposed here allows the use of other techniques in combination, and in some cases it would in fact benefit from the further development of those other techniques; for example, as noted before, if we learn to better specify what it means to impact the world as little as possible, that would help the approach proposed here, as this would allow us to better specify that the AI should just press the button and do nothing else.
We believe that the technique proposed in this paper, if properly further developed, may constitute one tool among others to help prevent uncontrolled powerful AI.

\section*{Acknowledgments}
I thank Lewis Hammond, Rubi Hudson, Vojta Kovarik, Chris van Merwijk, Sven Neth, and Caspar Oesterheld for helpful feedback.  I thank the Cooperative AI Foundation and Macroscopic Ventures (formerly Polaris Ventures
/ the Center for Emerging Risk Research) for financial support.

\bibliography{shutdown}
\bibliographystyle{plain}
\end{document}